\documentclass[letterpaper]{article} 
\usepackage{aaai2026}  
\usepackage{times}  
\usepackage{helvet}  
\usepackage{courier}  
\usepackage[hyphens]{url}  
\usepackage{graphicx} 
\usepackage{natbib}  
\usepackage{caption} 
\usepackage{multirow}
\usepackage{booktabs}
\usepackage{amsmath,amsfonts}
\usepackage{xcolor}

\usepackage{algorithm}
\usepackage{algorithmic}

\usepackage{newfloat}
\usepackage{listings}
\DeclareCaptionStyle{ruled}{labelfont=normalfont,labelsep=colon,strut=off} 
\floatstyle{ruled}
\newfloat{listing}{tb}{lst}{}
\floatname{listing}{Listing}
\title{Sim-to-Real: An Unsupervised Noise Layer for Screen-Camera Watermarking Robustness}
\author {
    Yufeng Wu\textsuperscript{\rm 1},
    Xin Liao\textsuperscript{\rm 1}\thanks{Corresponding author.},
    Baowei Wang\textsuperscript{\rm 2,3},
    Han Fang\textsuperscript{\rm 4},\\
    Xiaoshuai Wu\textsuperscript{\rm 1},
    Mingyue Chen\textsuperscript{\rm 1},
    Guiling Wang\textsuperscript{\rm 5}
}
\affiliations {
    \textsuperscript{\rm 1}College of Cyber Science and Technology, Hunan University, Changsha 410082, China\\
    \textsuperscript{\rm 2}the Engineering Research Center of Digital Forensics, Ministry of Education, the School of Computer Science,\\Jiangsu Collaborative Innovation Center of Atmospheric Environment and Equipment Technology,\\Nanjing University of Information Science and Technology, Nanjing 210044, China\\
    \textsuperscript{\rm 3}Jiangsu Yuchi Blockchain Technology Research Institute, Nanjing 210000, China\\
    \textsuperscript{\rm 4}School of Computing, National University of Singapore, Singapore\\
    \textsuperscript{\rm 5}Department of Computer Science, New Jersey Institute of Technology, Newark, NJ 07102, USA\\
    wuyufeng0523@163.com; xinliao@hnu.edu.cn; wang@nuist.edu.cn; fanghan@nus.edu.sg;\\ shinewu@hnu.edu.cn; chenmingyue@hnu.edu.cn; gwang@njit.edu
}

\usepackage{bibentry}

\begin{document}

\maketitle

\begin{abstract}
Unauthorized screen capturing and dissemination pose severe security threats such as data leakage and information theft. Several studies propose robust watermarking methods to track the copyright of Screen-Camera (SC) images, facilitating post-hoc certification against infringement. These techniques typically employ heuristic mathematical modeling or supervised neural network fitting as the noise layer, to enhance watermarking robustness against SC. However, both strategies cannot fundamentally achieve an effective approximation of SC noise. Mathematical simulation suffers from biased approximations due to the incomplete decomposition of the noise and the absence of interdependence among the noise components. Supervised networks require paired data to train the noise-fitting model, and it is difficult for the model to learn all the features of the noise. To address the above issues, we propose Simulation-to-Real (S2R). Specifically, an unsupervised noise layer employs unpaired data to learn the discrepancy between the modeled simulated noise distribution and the real-world SC noise distribution, rather than directly learning the mapping from sharp images to real-world images. Learning this transformation from simulation to reality is inherently simpler, as it primarily involves bridging the gap in noise distributions, instead of the complex task of reconstructing fine-grained image details. Extensive experimental results validate the efficacy of the proposed method, demonstrating superior watermark robustness and generalization compared to state-of-the-art methods.
\end{abstract}

\begin{links}
    \link{Code}{https://github.com/ttz0523/S2R-main}
\end{links}

\section{Introduction}
\begin{figure}[!htb]
	\centering
	\includegraphics[width=\linewidth]{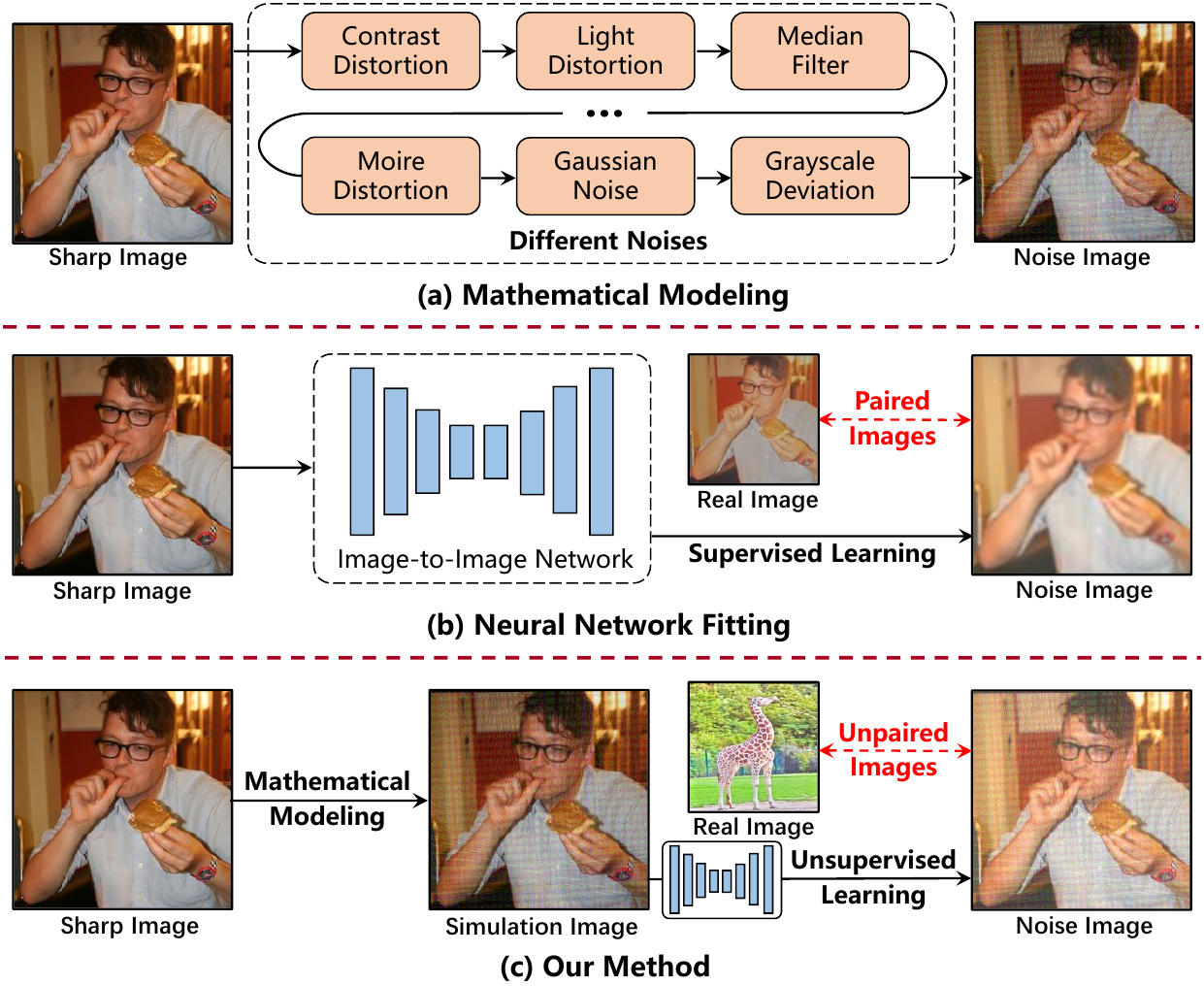}
	\caption{Overview of traditional noise approximation strategies and proposed Simulation-to-Real. (a) Mathematical modeling-based noise approximation. (b) Supervised neural fitting via paired images. (c) Our method: transforming sharp images to a certain noise domain, then mapping them unsupervised to an unknown domain, achieving more realistic noise approximation.
    }
	\label{fig:first}
\end{figure}
With the widespread use of digital images in areas such as digital photography, presentations, social media, and online publications, image piracy has become an increasing concern. Although digital watermarking techniques have proven effective for copyright protection in purely digital environments \cite{zhu2018hidden,jia2021mbrs,wang2023adaptor,fu2024waverecovery}, they frequently fail when content is presented on a screen and subsequently captured by a camera, which is a common method of unauthorized reproduction \cite{fang2018screen,tancik2020stegastamp}. In these Screen-Camera (SC) scenarios, watermarks are subject to complex physical degradations introduced during the display and capture process, which can lead to message loss. Ensuring robustness against SC degradation remains a critical challenge for watermarking systems in real-world copyright protection.

During the past few years, a large number of SC resistant watermarking methods have emerged. Adversarial training is an effective tool for resisting SC by approximating the noise introduced during SC. This approach helps to adapt the watermarking network to resist changes in the image caused by SC during training. To bridge the gap between digital and real environments, existing SC resistant watermarking works are categorized into two strategies: mathematical modeling \cite{tancik2020stegastamp,fang2022pimog,li2024screen} and neural network fitting \cite{wengrowski2019light}.

Mathematical modeling allows for a flexible and generalized SC noise approximation through a combination of different mathematical formulas, which explicitly expresses the impact of each noise component, as shown in Figure~\ref{fig:first}~(a). However, most mathematical modeling methods separate the noise components by linear superposition of independent noises \cite{tancik2020stegastamp,fang2022pimog,li2024screen}, ignoring the coupling of the noise sources in the real scene, leading to the deviation of the modeling from the real SC noise. Furthermore, mathematical modeling typically focuses on large-scale noise characteristics and struggles to model fine-grained, localized distortions. This is because the formulas and parameterizations employed in mathematical modeling are more suited to describing regular, widespread noise patterns, such as global blur, perspective distortions, or Gaussian noise.

Neural network fitting-based methods are capable of precisely fitting nonlinear SC noise features, as shown in Figure~\ref{fig:first}~(b). However, deep learning methods for noise approximation are hindered by the high demand for training data and limited noise modeling capacity. Current methods primarily adopt supervised learning, but obtaining diverse and high-quality paired SC samples is frequently a challenge. Constructing paired real samples typically involves manual rectification and alignment with sharp images, a process prone to spatial misalignments. Such inconsistencies introduce labeling bias, rendering data collection both error-prone and labor-intensive. Moreover, unlike mathematical models, directly mapping sharp images to the complex and variable noise patterns associated with SC can overwhelm the model due to the vast diversity of noise types, leading to suboptimal performance. Although the model is capable of generating precision noise, it may not capture detailed noise features, diminishing the precision of the approximation, making the training process more complex and hindering the convergence.

To address the aforementioned issues, in this work, we propose Simulation-to-Real (S2R), a novel framework that leverages mathematical modeling and neural network fitting to formulate the transition from sharp images to real-world SC noise. This work pioneers a neural network-based framework for fitting approximate noise, guided by mathematical modeling. As shown in Figure~\ref{fig:first}~(c), first, a rough representation of a certain domain is derived from a mathematical model. Then, the rough domain is further refined through unsupervised learning on unpaired data. This allows the model to capture unknown domain noise features that are not considered by the mathematical model.
We conduct extensive experiments to compare the effectiveness of our model with other state-of-the-art SC resistant watermarking methods. The results demonstrate that S2R outperforms existing methods, highlighting its superior performance in addressing the challenges of watermark robustness against SC in real-world environments.

The contributions of the proposed method are shown as follows:
\begin{itemize}
  \item We propose Simulation-to-Real (S2R), the first noise approximation framework guided by mathematical modeling. It maps simulated noise to real screen-camera noise and narrows the gap between synthetic and real data. The approach is supported by a feasibility proof.
  \item We innovatively introduce an unsupervised method without the requirement of paired data to bridge the distributional discrepancies between noise in simulation and real, resulting in improved accuracy in noise approximation.
  \item We build a scalable structure based on mathematical modeling and unsupervised learning. The mathematical modeling module supports replacement and flexible modification of algorithms according to requirements.
  \item Extensive experiments show that our method surpasses state-of-the-art methods in watermark robustness and image quality under real screen-camera conditions.
\end{itemize}
\begin{figure*}
    \centering
    \includegraphics[width=\linewidth]{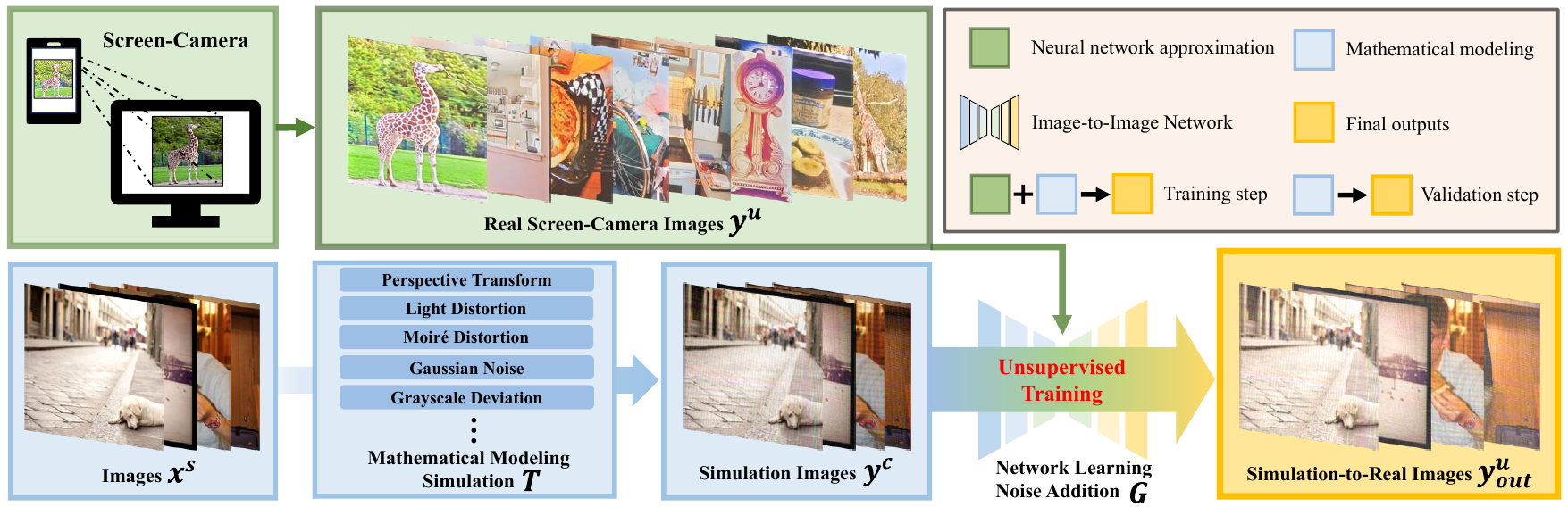}
    \caption{Overview of the proposed S2R. In the training phase, given a set of sharp images \( x^s \) and real SC images \( y^u \), the sharp images are first transformed into images with a certain simulation noise distribution \( y^c \) using a pre-defined mathematical modeling transformation \( T \). Through unsupervised training, the Image-to-Image Network \( G \) gradually adjusts \( y^c \) to match the distribution of \( y^u \), ultimately outputting the approximate images \( y_{\text{out}}^u \). In the validation phase, given sharp images \( x^s \), after passing through the transformation \( T \) and the fixed-weight network \( G \), the outputs are \( y_{\text{out}}^u \).}
    \label{fig:S2Rframework}
    
\end{figure*}
\section{Related Work}

\subsection{Screen-Camera Resistant Watermarking}

Recent SC resistent watermarking methods increasingly adopt deep learning frameworks \cite{zhu2018hidden,wang2023adaptor} due to their superior robustness and imperceptibility compared to traditional techniques \cite{fang2020deep,10321686,wang2024robust,zhu2024lite}. These methods typically use an encoder–decoder structure, with a noise layer inserted between them to simulate distortions during training. The noise layer designed specifically to simulate SC noise can be constructed either through mathematical modeling or neural network fitting.

Mathematical modeling-based strategy is widely adopted in watermarking frameworks targeting SC distortions, due to its differentiability and controllable approximation capability. StegaStamp \cite{tancik2020stegastamp} proposed a differentiable pipeline to simulate physical distortions such as perspective distortion, blur, color shifts, noise, and JPEG compression. Unlike earlier single-noise strategies, it applies all distortions sequentially to improve robustness. Luo et al. \cite{luo2020distortion} added lighting variations to enhance realism. PIMoG \cite{fang2022pimog} modeled SC noise as a mixture of perspective transform, lighting, moiré, and Gaussian noise. SSDS \cite{li2024screen} further introduced grayscale deviation. Although effective, these models still struggle to capture complex artifacts in real SC scenarios.

Neural network fitting-based strategy learn distortion mappings directly from data. \cite{wengrowski2019light} trained a distortion network, CDTF, with a 1.9 TB real dataset, allowing the network to learn the features of SC distortion. Similarly, CDTF has also been successfully applied in challenging Print-Camera  scenarios \cite{qin2023print}. However, these supervised methods rely heavily on paired data and frequently generalize poorly to unseen devices, making them less suitable in real-world settings. Thus, the neural network fitting-based strategy requires further exploration.

\subsection{Unsupervised Learning with Unpaired Data}

To overcome the dependence on paired data, unsupervised learning methods based on Generative Adversarial Networks (GANs) \cite{goodfellow2014generative} have been extensively studied. In particular, CycleGAN \cite{zhu2017unpaired} and DualGAN \cite{yi2017dualgan} introduce cycle consistency loss to enable unpaired image translation, facilitating learning from unpaired data. While Pix2Pix \cite{isola2017image} represents a supervised image-to-image translation framework relying on paired data and adversarial loss with a PatchGAN discriminator, its core principles have been extended to unsupervised settings \cite{pham2024blur2blur}. This has led to progress in deblurring \cite{pham2024blur2blur}, deraining \cite{chang2023unsupervised}, and denoising \cite{pang2021recorrupted}.

However, these frameworks struggle with bridging the gap between sharp and noisy domains due to the complexity of real-world noise. To address this limitation, our S2R builds upon employing unsupervised techniques to learn and approximate the differences between simulated noise distributions and real SC noise. This approach overcomes the challenge of direct mapping between simulated and real noise complexities.

\section{Proposed Method}
\subsection{Motivation}

Our goal is to learn a noise approximation function \( F_\mathcal{U}(\cdot) \) that transforms sharp images \( x^s \in \mathcal{S} \) into SC images \( y^u \in \mathcal{U} \), i.e., \( F_\mathcal{U}(x^s) = y^u \).

One strategy for approximating SC noise is to decompose it into independent components via mathematical modeling, thereby enabling targeted simulation. However, this strategy has two significant limitations. First, it tends to introduce superfluous or overly pronounced distortions, which compromises imperceptibility. Second, by assuming component independence, it disregards the inherent interdependencies among different noise types, leading to unrealistic approximations of real-world conditions.

An alternative approach is to employ supervised image translation networks to directly learn the mapping \( F_\mathcal{U}(\cdot) \) from paired data. However, these methods face two main limitations. First, obtaining reliable supervised data is inherently challenging. SC images require manual rectification and cropping to align with their original counterparts, a process that often introduces spatial misalignments. These misalignments lead to significant labeling bias, making the collection of truly accurate data pairs both costly and labor-intensive. Second, even with ideal data, neural networks with limited capacity struggle to model the highly complex and diverse nature of SC noise. The underlying noise distribution often comprises a mixture of intricate, overlapping patterns that exceed the representational ability of a single network. As a result, the network fails to capture the full diversity and complexity of real-world noise variations.

\subsection{Method Overview}
Based on the above discussion, when neither mathematical modeling nor neural network fitting is ideal due to their inherent limitations, a hybrid strategy that leverages the strengths of both becomes a promising alternative for approximating SC noise. We introduce an innovative method of learning \( F_\mathcal{U}(\cdot) \), rather than directly learning this function, which is very challenging. We treat \( F_\mathcal{U}(\cdot) \) as a composition of the mathematical modeling transformation \( T \) from the certain domain \( \mathcal{C} \) and the network mapping function \( G \) from the certain domain to the unknown domain \( \mathcal{U} \): 
\begin{equation}
	F_\mathcal{U}(\cdot) = T * G.
\end{equation}

We leverage the prior noise constraints provided by mathematical modeling to offer a foundation of noise robustness and generalization for deep learning, and then utilize deep learning to bridge the gap between the certain modeling noise distribution and the unknown noise distribution. The challenge here is that we cannot obtain paired noise datasets for supervised training, as acquiring paired datasets of simulated images and real SC images is extremely difficult. Therefore, the only feasible choice is to adopt unpaired data. Fortunately, we can capture a set of real SC images \( U \) arbitrarily, without being restricted by the set of sharp images \( S \). These two sets of images are unpaired, which means that there is no need for a one-to-one correspondence between images from \( \mathcal{S} \) and \( \mathcal{U} \). Consequently, collecting these datasets is relatively easy and straightforward.

 Our goal is then shifted to learning \( G \) to bridge the gap between domain \( \mathcal{C} \) and domain \( \mathcal{U} \). In particular, our task is to learn a mapping function \( G \) that maps each input image \( y^c \) simulated by mathematical modeling, defined in Eq.~(\ref{overview}), to an image \( y^u \) that has the same sharp visual representation \( x^s \) but belongs to the unknown noise distribution \( \mathcal{U} \).
\begin{equation}
	\label{overview}
	T : x^s \to y^c, G : y^c \to y^u, \quad \text{where} \quad y^u = G(T(x^s)).
\end{equation}

The general method is illustrated in Figure~\ref{fig:S2Rframework}. Our method decomposes a complex task into two more manageable tasks. One task is to simulate noise through mathematical modeling, which, although challenging, benefits from existing research. We can select a well-performing noise model \( T \) that has already achieved good simulation results in its domain \( \mathcal{C} \). The other task is to learn the transformation from a certain noise domain \( \mathcal{C} \) to the unknown domain \( \mathcal{U} \). The difficulty of this task depends on the difference between \( \mathcal{C} \) and \( \mathcal{U} \), but it is much easier than directly learning the mapping from \( \mathcal{S} \) to \( \mathcal{U} \). Moreover, we can flexibly select the most appropriate \( T \) and \( G \) for our specific SC noise domain, ensuring that the simulation accuracy of SC noise is maximized.

\textbf{Design of S2R.}
To implement S2R framework, the key is to train a noise-to-noise transformation network \( G \) that can convert any noisy image from a certain noise domain \( \mathcal{C} \) to an unknown noise domain \( \mathcal{U} \), while preserving the content of the image. To train \( G \), we need two datasets: sharp images from \( \mathcal{S} \), and real SC images from an unknown source \( \mathcal{U} \). We design \( G \) to operate on multiple scales and carefully design the training loss to achieve the desired results.
We adopt MIMO-UNet \cite{fan2021rethinking} for its efficiency and simplicity. For a more detailed description of the framework, please refer to Section A.3 of the supplementary material.

\begin{figure}[!htb]
	\centering
	\includegraphics[width=\linewidth]{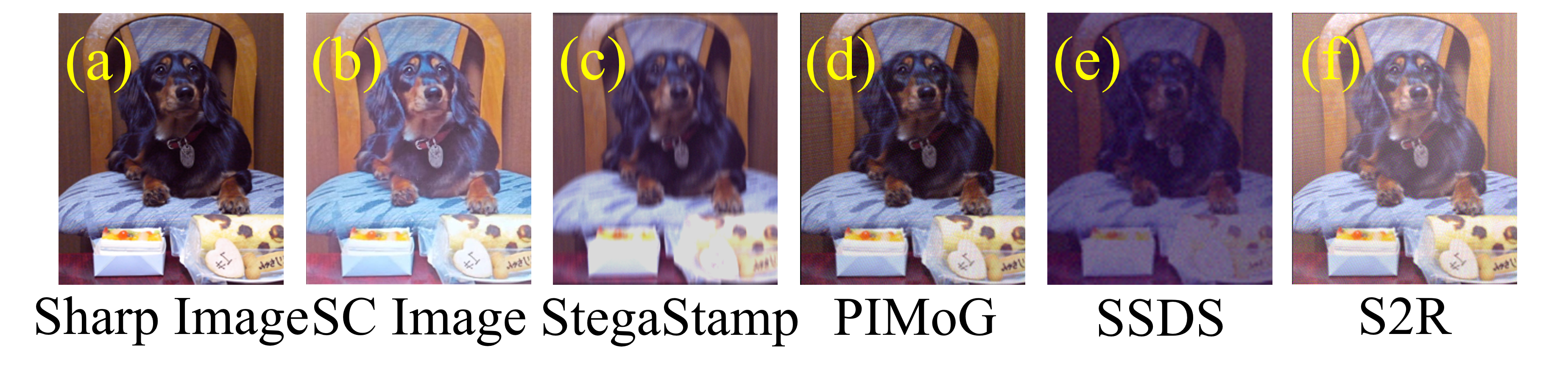}
	\caption{Approximation results of different methods converting sharp images into noise images: (a) sharp images; (b) real SC images; (c) StegaStamp \cite{tancik2020stegastamp}; (d) PIMoG \cite{fang2022pimog}; (e) SSDS \cite{li2024screen}; (f) our S2R.
    }
	\label{fig:noiseex}
    
\end{figure}

\subsection{Loss Function}
\subsubsection{Adversarial Loss}
We employ adversarial loss to constrain the generative network \( G \) to produce images with the characteristics of target noise. To achieve this, we introduce a discriminator network \( D \) to distinguish between real SC noisy images and generated images. Networks \( G \) and \( D \) are alternately trained within a minimax game framework. The adversarial loss is defined as follows:
\begin{equation}
	L_{\text{cGAN}}(G, D) = \mathbb{E}_{y \sim \mathcal{U}} \left[ \log D(y) \right] + \mathbb{E}_{y \sim \mathcal{C}} \left[ \log \left( 1 - D(G(y)) \right) \right].
\end{equation}
We train \( G \) to minimize the above loss term, while training the discriminator \( D \) to maximize it. Additionally, we apply the regularization of gradient penalty on the discriminator to enforce the Lipschitz continuity constraint \cite{gulrajani2017improved}. The gradient penalty loss is defined as:
\begin{equation}
	L_{\text{grad}}^D (D) = \mathbb{E}_{\tilde{y} \sim \tilde{C}} \left[ \left( \left\| \nabla_{\tilde{y}} D(\tilde{y}) \right\|_2 - 1 \right)^2 \right],
\end{equation}
where \( \tilde{C} \) represents the set of samples \( \tilde{y} \) obtained through random interpolation between an image \( y \in C \) and the generated image \( G(y) \).

\subsubsection{Reconstruction Loss}
Previous methods \cite{zhu2017unpaired} have found it beneficial to combine the GAN objective with more traditional losses (such as the $L_1$, $L_2$ distance, etc.). The discriminator's task remains unchanged, but the generator's objective is to both deceive the discriminator and generate outputs consistent with the real noise. We adopt a weighted multi-scale perceptual loss which reconstructs image content from coarse to fine without being overly constrained by pixel-level accuracy:
\begin{equation}
	L_P (G) = \frac{1}{2^{k-1}} \sum_{i=1}^{k} \frac{1}{t_i} \mathbb{E}_{y \sim \mathcal{U}} \left\| \varphi(y_i^c) - \varphi(G(y_i^c)) \right\|,
\end{equation}
where $\varphi$ is a feature extractor from a pretrained network.
\subsubsection{Total Loss}
By combining \( L_{\text{cGAN}} \) with \( L_P \), the generator loss \( L_G \) and the discriminator loss \( L_D \) are given as follows:
\begin{equation}
	L_G = L_{\text{cGAN}}(G, D) + \lambda_G L_P(G),
\end{equation}
\begin{equation}
	L_D = -L_{\text{cGAN}}(G, D) + \lambda_{\text{grad}} L_{\text{grad}}^D(D),
\end{equation}
where \( \lambda_G \) and \( \lambda_{\text{grad}} \) are the weight factors.

\subsection{Feasibility Proof of Inter-domain Migration for Noise}
In this work, to address the difficulty of approximating SC noise, our method transforms from \( \mathcal{S} \) to \( \mathcal{C} \), and then estimates the unknown real SC noise domain \( \mathcal{U} \), i.e., \( \mathcal{S} \to \mathcal{C} \to \mathcal{U} \). We focus on the noise in the SC input, treating the added noise as multiplicative and additive noise \cite{lim1990two}. The noise image \( y^u \) is denoted as a function of the corresponding sharp image \( x^s \) through the noise operators \( k^u \) and \( n^u \), with these operators being associated with the noise domain \( \mathcal{U} \) that corresponds to the SC noise:
\begin{equation}
	y^u = k^u \cdot x^s + n^u,
\end{equation}
where \( k^u \) is the multiplicative noise term, and \( n^u \) is the additive noise term. Then, we decompose the new unknown noise formula into the following operators:

(a) Multiplicative noise term:
\begin{equation}
	k^u = k^{(c \rightarrow u)} \cdot k^{(s \rightarrow c)}.
\end{equation}

(b) Additive noise term:
\begin{equation}
	\begin{split}
		n^u &= k^{(c \rightarrow u)} \cdot n^c + n^{(c \rightarrow u)} \\
		&= k^{(c \rightarrow u)} \cdot \left(k^{(s \rightarrow c)} \cdot n^s + n^{(s \rightarrow c)}\right) + n^{(c \rightarrow u)},
	\end{split}
\end{equation}
where \( k^{(c \rightarrow u)} \) denotes the degree to which \( k^u \) deviates from any noise \( k^c \) sampled from domain \( \mathcal{C} \) (multiplicative noise operator), and \( n^{(c \rightarrow u)} \) denotes the degree to which \( n^u \) deviates from any noise \( n^c \) sampled from domain \( \mathcal{C} \) (additive noise operator). Similarly, other operators can be derived. Given the above formulas, we observe that the multiplicative noise \( k^u \) can be decomposed as \( k^{(c \rightarrow u)} \cdot k^{(s \rightarrow c)} \), and the additive noise can be decomposed as:
\begin{equation}
	n^u = k^{(c \rightarrow u)} \cdot \left(k^{(s \rightarrow c)} \cdot n^s + n^{(s \rightarrow c)}\right) + n^{(c \rightarrow u)}.
\end{equation}
Therefore, we can express the noise image \( y^u \) as follows:
\begin{equation}
	\begin{aligned}
		y^u &= k^u \cdot x^s + n^u \\
		&= k^{(c \rightarrow u)} \cdot k^{(s \rightarrow c)} \cdot x^s + k^{(c \rightarrow u)} \cdot \left(k^{(s \rightarrow c)} \cdot n^s + n^{(s \rightarrow c)}\right) \\
		&\quad + n^{(c \rightarrow u)}.
	\end{aligned}
\end{equation}
Further expansion:
\begin{equation}
	\begin{aligned}
		y^u&= k^{(c \rightarrow u)} \cdot k^{(s \rightarrow c)} \cdot x^s + k^{(c \rightarrow u)} \cdot k^{(s \rightarrow c)} \cdot n^s \\
		&\quad + k^{(c \rightarrow u)} \cdot n^{(s \rightarrow c)} + n^{(c \rightarrow u)}.
	\end{aligned}
\end{equation}
Factor out the common term \( k^{(c \rightarrow u)} \):
\begin{equation}
	\begin{aligned}
		y^u&= k^{(c \rightarrow u)} \cdot \left( k^{(s \rightarrow c)} \cdot x^s + n^{(s \rightarrow c)} \right) \\
		&\quad + n^{(c \rightarrow u)} + k^{(c \rightarrow u)} \cdot k^{(s \rightarrow c)} \cdot n^s.
	\end{aligned}
\end{equation}
Since \( y^c = k^{(s \rightarrow c)} \cdot x^s + n^{(s \rightarrow c)} \), substituting this gives:
\begin{equation}
	y^u= k^{(c \rightarrow u)} \cdot y^c + n^{(c \rightarrow u)} + k^{(c \rightarrow u)} \cdot k^{(s \rightarrow c)} \cdot n^s.
\end{equation}
In our method, since all the sharp images are assumed to be noise-free, we can set \( n^s = 0 \). The equation simplifies to:
\begin{equation}
	y^u= k^{(c \rightarrow u)} \cdot y^c + n^{(c \rightarrow u)}.
\end{equation}
We convert \( k^{(c \rightarrow u)} \) and \( n^{(c \rightarrow u)} \) into the corresponding neural network mappings \( k_\delta \) and \( n_\delta \), respectively. Ultimately, we can conclude:
\begin{equation}
	y^u= k_\delta \cdot y^c + n_\delta.
\end{equation}

Through the above derivation, the distribution alignment task is transformed from directly learning the sharp image \( x^s \) to \( y^u \) into learning the bias between the certain noise image \( y^c \) and the unknown noise image \( y^u \). This significantly reduces the difficulty of the unsupervised distribution alignment task and facilitates the network's ability to correctly focus on learning the differences between domain \( \mathcal{C} \) and domain \( \mathcal{U} \). Therefore, we are able to utilize neural networks to accomplish the conversion from the certain noise domain to the unknown noise domain to obtain an accurate simulation of the real SC noise.

\section{Experimentation}
\subsection{Implementation Details}

\subsubsection{Watermarking and S2R Framework}

We adopt MCFN \cite{WU2024129282} as the default watermarking framework and employ the COCO dataset \cite{lin2014microsoft} for training. Following previous works \cite{jia2021mbrs,fang2022pimog,wang2023adaptor}, 10{,}000 images are selected for training, each resized to $128 \times 128$ and embedded with a random 64-bit binary watermark.
To simulate SC degradation, we propose the S2R framework, which consists of a simulated noise layer \( T \) and a transformation network \( G \). We use PIMoG \cite{fang2022pimog} as \( T \), and an improved MIMO-UNet \cite{fan2021rethinking} as \( G \) with the default configuration in unsupervised Pix2Pix implementation \cite{isola2017image, pham2024blur2blur}. To build the training dataset for S2R, we randomly select COCO images and capture their SC versions by three device pairs: Samsung Galaxy S20 FE with Lenovo Legion Y9000P (S+L), iPhone 13 with Envision G249G (I+E), and MEIZU 20 Pro with ASUS ROG Strix SCAR Edition 8 (M+A). Each pair contributes 900 SC images, forming a combined dataset referred to as SIM+LEA. All training is conducted on an NVIDIA RTX 4090 GPU.

For default testing, we train S2R on the SIM+LEA dataset. We then capture 100 encoded images from the COCO dataset, which was not used during training, with the S+L pair for evaluation. Viewpoint angles follow a left-to-right axis, where negative and positive values indicate bottom/left and top/right tilts, respectively. Comparative methods are tested under the same conditions to ensure fairness.
More details are provided in Section A of the supplementary material.

\begin{figure*}[!htb]
	\centering
	\includegraphics[width=\linewidth]{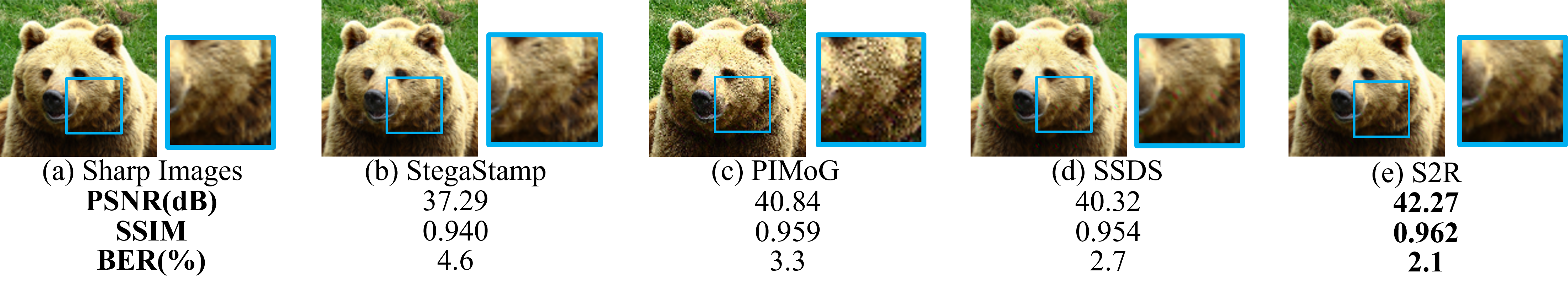}
	\caption{Visual quality and robustness comparison of watermarking methods: (a) Original, (b) StegaStamp \cite{tancik2020stegastamp}, (c) PIMoG \cite{fang2022pimog}, (d) SSDS \cite{li2024screen}, (e) Proposed S2R (trained on SIM+LEA).
	}
	\label{fig:visualex}
\end{figure*}

\subsubsection{Metrics}
We evaluate watermarking performance in terms of invisibility and robustness. Peak Signal-to-Noise Ratio (PSNR) and Structural Similarity Index Measure (SSIM) measure visual quality between the watermarked and original images, where higher values indicate less distortion. Robustness is assessed via the Bit Error Rate (BER), with lower values indicating more accurate extraction.

\subsubsection{Baseline}
To ensure fair comparison, we benchmark S2R against StegaStamp \cite{tancik2020stegastamp}, PIMoG \cite{fang2022pimog}, and SSDS \cite{li2024screen}. Since SSDS and CDTF \cite{wengrowski2019light} lack open-source implementations, we reproduce SSDS and replace CDTF with a supervised variant of S2R.

\begin{table}[!t]

\begin{tabular}{@{}cccccccc@{}}
\toprule
\multicolumn{2}{c}{\multirow{2}{*}{Distance = 30 cm}} & \multicolumn{6}{c}{Training dataset}                                              \\ \cmidrule(l){3-8} 
\multicolumn{2}{c}{}                                  & \multicolumn{2}{c}{S+L}   & \multicolumn{2}{c}{I+E}   & \multicolumn{2}{c}{M+A}   \\ \midrule
\multicolumn{2}{c}{PSNR (dB)}                              & \multicolumn{2}{c}{41.94} & \multicolumn{2}{c}{41.00} & \multicolumn{2}{c}{42.57} \\ \midrule
\multicolumn{2}{c}{SSIM}                              & \multicolumn{2}{c}{0.969} & \multicolumn{2}{c}{0.957} & \multicolumn{2}{c}{0.964} \\ \midrule
\multicolumn{2}{c}{Angle}                             & 0°          & 40°         & 0°           & 40°        & 0°          & 40°         \\ \midrule
\multirow{3}{*}{Device pair}           & S+L          & 2.6         & 6.1         & 2.2          & 4.9        & 1.6         & 5.1         \\
                                       & I+E          & 2.1         & 5.0         & 2.5          & 4.5        & 2.2         & 5.6         \\
                                       & M+A          & 2.2         & 5.3         & 2.3          & 5.3        & 2.0         & 5.9         \\ \bottomrule
\end{tabular}
\caption{Generalization Robustness Test Results of S2R Trained on Datasets with Different Device Pairs.\label{tab:genrobust}}
\end{table}

\subsection{Noise Approximation Experiments}

In noise approximation experiments, we visually compare the output of different methods and analyze the noise differences. As shown in Figure~\ref{fig:noiseex}, other methods exhibit limitations in detail and noise realism. StegaStamp produces darker images because of lower saturation, brightness, and contrast. PIMoG improves visually but struggles with lightness variations. SSDS shows improvements but still suffers from darkening. In contrast, S2R generates more natural images, closely resembling real SC noise and bridging the gap between physical and approximate environments. Histogram comparisons further confirm S2R's superior approximation of subtle noise variations, detailed in Section C of the supplementary material.
These results confirm that S2R outperforms existing modeling methods in both perceptual fidelity and statistical consistency, offering a more accurate and realistic noise layer.

\begin{table}[!t]
	\centering

        \setlength{\tabcolsep}{1mm}
		\begin{tabular}{cccccc}
		\toprule
			Distance = 30 cm & \multicolumn{2}{c}{Image quality} & \multicolumn{3}{c}{BER (\%)} \\ \cmidrule(l){2-6}
			Method      & PSNR (dB)          & SSIM           & 0°       & 20°      & 40°      \\ \midrule
            StegaStamp       & 39.89            & 0.948          & 5.5     & 7.1     & 7.3     \\
			PIMoG       & 41.41            & 0.950          & 6.2     & 8.8     & 9.5     \\
			SSDS          & 41.05            & 0.956          & 5.1     & 6.0     & 7.6     \\
			S2R         & \textbf{42.27}            & \textbf{0.962}          & \textbf{2.1}     & \textbf{3.3}     & \textbf{6.0}     \\ \bottomrule
	\end{tabular}
        \caption{Performance Comparison of Models with Different Noise Layers under the Same Watermarking Framework.\label{tab:difnoiselayer}}
\end{table}

\begin{table*}[!t]
\centering
\setlength{\tabcolsep}{1.5mm}
\begin{tabular}{@{}cccccccclccccccccccc@{}}
\toprule
BER (\%) & \multicolumn{5}{c}{Angle = 0°} & \multicolumn{7}{c}{Distance = 30 cm (Left to Right)} & \multicolumn{7}{c}{Distance = 30 cm (Up to Down)} \\ 
\cmidrule(lr){1-1} \cmidrule(lr){2-6} \cmidrule(lr){7-13} \cmidrule(lr){14-20}
Methods & 20cm & 25cm & 30cm & 35cm & 40cm & -60° & -40° & -20° & 20° & 40° & \multicolumn{2}{c}{60°} & \multicolumn{2}{c}{-60°} & -40° & -20° & 20° & 40° & 60° \\ 
\midrule
StegaStamp & 2.9 & 3.9 & 4.6 & 4.7 & 4.4 & 5.9 & 7.2 & 4.1 & 5.8 & 7.7 & \multicolumn{2}{c}{7.6} & \multicolumn{2}{c}{15.1} & 4.8 & 4.7 & 3.5 & 5.7 & 9.4 \\
PIMoG      & 1.5 & 1.4 & 3.3 & 3.2 & 2.6 & 9.0 & 8.7 & 5.2 & 5.3 & 9.3 & \multicolumn{2}{c}{9.7} & \multicolumn{2}{c}{15.3} & 4.7 & 4.5 & 2.1 & 5.0 & 6.5 \\
SSDS       & 2.4 & 2.7 & 2.1 & 2.7 & 4.1 & 7.5 & 5.1 & 3.9 & 4.2 & 6.1 & \multicolumn{2}{c}{6.2} & \multicolumn{2}{c}{9.3}  & 4.0 & 3.3 & 2.9 & 4.8 & 6.2 \\
S2R (SIM+LEA) & \textbf{1.2} & \textbf{1.1} & \textbf{2.1} & \textbf{2.5} & \textbf{2.2} & \textbf{5.8} & \textbf{3.9} & \textbf{3.2} & \textbf{3.3} & \textbf{6.0} & \multicolumn{2}{c}{\textbf{5.9}} & \multicolumn{2}{c}{10.2} & \textbf{3.3} & \textbf{2.3} & \textbf{1.8} & \textbf{4.2} & 6.3 \\
\bottomrule
\end{tabular}
\caption{Comparison of Bit Accuracy for Extracted Watermark Message under Different Capture Distances and Angles.\label{tab:angle}}
\end{table*}

\subsection{Ablation Experiments}
\subsubsection{Cross-Device Generalization}
To validate the generalization of S2R, we conduct a cross-device generalization test. Specifically, we train the model on datasets from three device pairs (S+L, I+E, M+A) and then test the trained S2R on other device pairs.
In the experiment, the shooting conditions are consistent with two capture angles (0° and 40° from the left) for comparison, and a shooting distance of 30 cm is maintained. The results of the generalization experiments, as shown in Table~\ref{tab:genrobust}, indicate that despite hardware differences between devices, the S2R is still able to effectively extract and recover the watermark, demonstrating high accuracy between datasets from different devices.

\subsubsection{Cross-Dataset Generalization}
We further evaluate the generalization of the framework generalization by testing on datasets unseen during watermarking training. Despite variations in image semantics and textures, the method maintains similar robustness and perceptual quality as on the original dataset, demonstrating strong adaptability under domain shifts. This robustness mainly arises from the model learning SC noise characteristics rather than content-specific features. Detailed results are reported in Section D.1 of the supplementary material.

\subsubsection{Cross-Source Training for S2R and Watermarking}
To further test the generalization, we train the noise generation module and watermarking network on different datasets. Even with this cross-source training, the system maintains low BER and preserves image quality, with only minor perceptual quality drops due to source differences. These findings confirm that our framework supports flexible, decoupled training of its components without sacrificing robustness. Additional analyses are available in Section D.2 of the supplementary material.

\subsubsection{Comparison of Different Noise Layers Under the Same Watermarking Framework}

To validate the effectiveness of the S2R noise layer, we train multiple models under an identical watermarking framework, MCFN \cite{WU2024129282}, with different noise layers. We select different noise layers from StegaStamp, PIMoG, and the state-of-the-art SSDS, which are widely adopted in contemporary SC resistant watermarking schemes, as baselines, and compare their performance with S2R.
As shown in Table~\ref{tab:difnoiselayer}, S2R outperforms other methods in all aspects. This is because the core advantage of S2R lies in the ability to adaptively adjust the noise distribution. The model can rely on predefined mathematical modeling of noise and bridge the noise distribution based on the statistical characteristics of the data. Therefore, S2R can more accurately approximate the noise distribution in real SC, surpassing traditional noise layer designs.

\subsection{Comparison Experiments}
\subsubsection{Comparison of Visual Quality}
In this experiment, we compare the invisibility of watermark images generated by different methods, with the results presented in Figure~\ref{fig:visualex}. Compared to other methods, S2R reduces visible artifacts due to the superior design of its noise layer, which bridges the gap between mathematical modeling and real-world distortions. This design enables S2R to better approximate the statistical properties and intensity distribution of real SC noise, allowing the adversarial training process to be more closely aligned with real scenes. Consequently, the model requires less effort to counteract extraneous noise, enhancing its visual quality in real SC scenarios.

\subsubsection{Comparison of Robustness Under Different Shooting Distances}
To assess robustness against variations in shooting distance, we conduct experiments across five distances ranging from 20 cm to 40 cm under fixed perpendicular capture angles. As shown in Table~\ref{tab:angle}, S2R consistently achieves the lowest BER on all distances, outperforming existing methods. This robustness is primarily due to S2R’s ability to handle resolution degradation and potential defocus effects introduced by distance changes. By combining mathematical modeling simulation and unsupervised refinement, S2R learns to adapt to these changes and maintain reliable watermark recovery in real SC scenarios.

\subsubsection{Comparison of Robustness Under Different Shooting Angles}
We conduct experiments by capturing watermark images at angles ranging from 0° to 60°, both from left to right and from bottom to top, and compare the BER. The experimental results in Table~\ref{tab:angle} show that within smaller angle ranges, the model is able to recover the watermark effectively, with the BER remaining at a low level. At larger shooting angles, the degradation in watermarked image quality results in increased BER. However, S2R consistently achieves lower error rates than competing approaches, indicating stronger robustness to angle-induced distortions.

\begin{table}[!t]
	\centering
        \setlength{\tabcolsep}{1mm}
		\begin{tabular}{cccccc}
			\toprule
			Distance = 30 cm    & \multicolumn{2}{c}{Image quality} & \multicolumn{3}{c}{BER (\%)} \\ \midrule
			Model         & PSNR          & SSIM           & 0°       & 20°      & 40°      \\ \midrule
			StegaStamp-based(SIM+LEA)    & 40.47            & 0.952          & 2.4     & 3.7     & 7.1     \\
			SSDS-based(SIM+LEA)       &  41.25                &  0.967              & 5.0        & 8.1        &  10.6    \\ \midrule
			S2R-supervised(I+E) &  41.29                & 0.959               & 3.8       & 5.5     &  7.9   \\
            S2R-CycleGAN(SIM+LEA) &  41.85                & 0.960               & 2.9       & 4.5     &  6.9   \\
            S2R-DualGAN(SIM+LEA) &  41.55                & 0.958               & 3.5       & 5.2     &  7.6   \\ \midrule
			S2R(I+E)      & \textbf{42.57}            & 0.964          & \textbf{1.6}     & \textbf{3.1}     & \textbf{5.1}     \\
			S2R(SIM+LEA)  & 42.27            & 0.962          & 2.1     & 3.3     & 6.0     \\ \bottomrule
	\end{tabular}
        \caption{Performance Comparison of Models with Different Mathematical Modeling Methods and Supervision Methods.\label{tab:noiseandsupervised}}
\end{table}

\subsection{Scalability Experiments}

To assess the scalability of S2R and compare different noise modeling strategies, we integrate noise layers from StegaStamp, SSDS, and PIMoG into the unsupervised S2R framework. These variants, denoted as StegaStamp-based, SSDS-based, and PIMoG-based (S2R), demonstrate that the framework supports plug-and-play replacement of degradation models without altering its core structure. To further assess S2R, we train a supervised variant of S2R to learn a direct mapping from sharp to noisy images based on 900 paired samples from I+E. As shown in Table~\ref{tab:noiseandsupervised}, although the S2R-supervised benefits from paired supervision, its performance remains limited due to insufficient generalization and reliance on labeled data. 
In addition, we implement DualGAN and CycleGAN as alternative unsupervised methods. Although they underperform compared to S2R, both methods outperform counterparts that do not employ unsupervised networks for noise modeling. This highlights the advantage of incorporating learning-based strategies over purely mathematical modeling. In summary, S2R achieves greater robustness, imperceptibility, and generalization.

\section{Conclusion}
In this paper, we propose a novel SC-resistant watermarking framework, Simulation-to-Real (S2R), which improves robustness against SC noise in real-world scenarios. S2R adopts a two-step strategy: it first simulates rough noise via mathematical modeling, then refines them with an unsupervised image-to-image network to approximate real SC noise. Experiments show that S2R achieves superior robustness and generalization compared to existing methods.
In future work, we will enhance flexibility by enabling adaptive noise refining during end-to-end training.

\section{Acknowledgments}
This work is supported by National Key R\&D Program of China (Grant Nos. 2024YFF0618800, 2022YFB3103500), National Natural Science Foundation of China (Grant Nos. U22A2030, U22B2062), Hunan Provincial
Funds for Distinguished Young Scholars (Grant No. 2024JJ2025), Hunan Provincial Key Research and Development Program (Grant Nos. 2024AQ2027, 2025AQ2022), Nanjing Major Science and Technology Special Project (Grant No. 202405002).

\bibliography{aaai2026}

\setlength{\leftmargini}{20pt}
\makeatletter\def\@listi{\leftmargin\leftmargini \topsep .5em \parsep .5em \itemsep .5em}
\def\@listii{\leftmargin\leftmarginii \labelwidth\leftmarginii \advance\labelwidth-\labelsep \topsep .4em \parsep .4em \itemsep .4em}
\def\@listiii{\leftmargin\leftmarginiii \labelwidth\leftmarginiii \advance\labelwidth-\labelsep \topsep .4em \parsep .4em \itemsep .4em}\makeatother

\setcounter{secnumdepth}{0}
\renewcommand\thesubsection{\arabic{subsection}}
\renewcommand\labelenumi{\thesubsection.\arabic{enumi}}

\newcounter{checksubsection}
\newcounter{checkitem}[checksubsection]

\newcommand{\checksubsection}[1]{%
  \refstepcounter{checksubsection}%
  \paragraph{\arabic{checksubsection}. #1}%
  \setcounter{checkitem}{0}%
}

\newcommand{\checkitem}{%
  \refstepcounter{checkitem}%
  \item[\arabic{checksubsection}.\arabic{checkitem}.]%
}
\newcommand{\question}[2]{\normalcolor\checkitem #1 #2 \color{blue}}
\newcommand{\ifyespoints}[1]{\makebox[0pt][l]{\hspace{-15pt}\normalcolor #1}}

\clearpage
\appendix
\section{Supplementary Material}

This supplementary material provides additional implementation details,
experimental results, and visualizations supporting the main paper.

\begin{figure*}
	\centering
	\includegraphics[width=\linewidth]{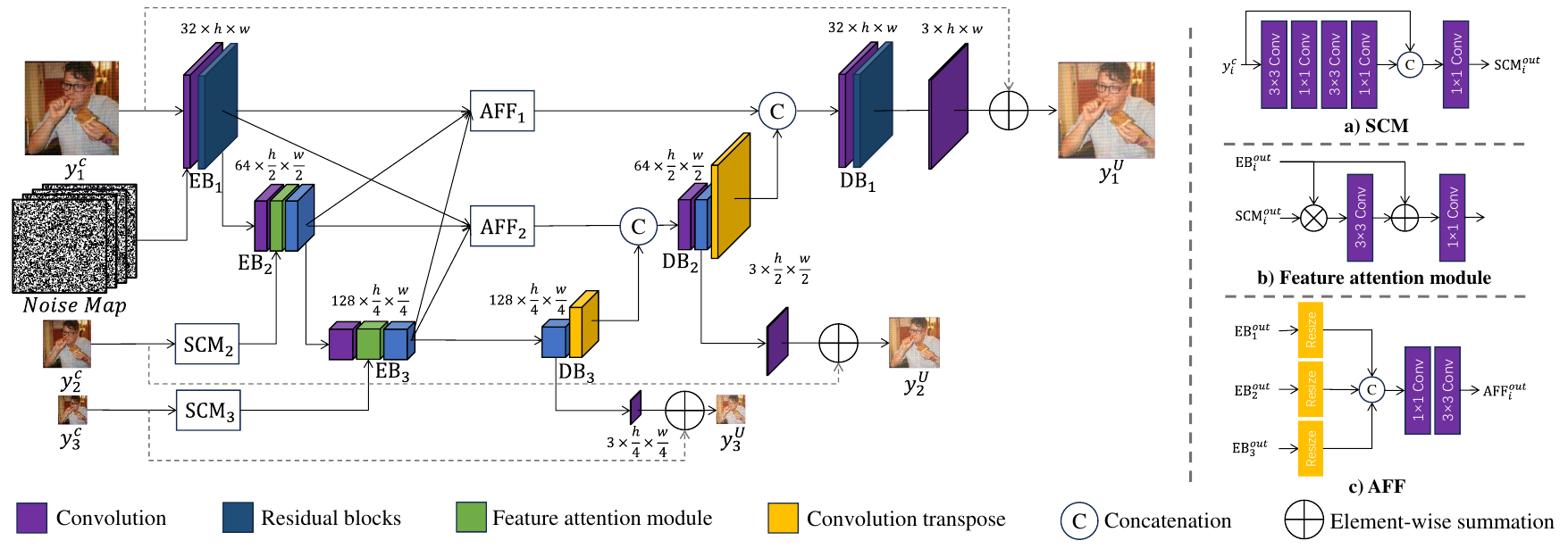}
	\caption{The overall framework of the modified MIMO-UNet. It consists of an encoder that receives downsampled noise images at different scales and a Gaussian noise map sampled from a random normal distribution as input. The Shallow Convolutional Module (SCM) initially extracts features, followed by an Encoder Block (EB) containing a Feature Attention Module (FAM) that progressively extracts multi-level features. Between the encoder and decoder, an Asymmetric Feature Fusion (AFF) module is introduced to effectively fuse features at different scales. During the decoding phase, the network uses a single decoder to generate multiple deblurred images at different scales via the Decoder Block (DB). On the right side of the figure, from top to bottom, are the structures of the SCM, Feature Attention Module and AFF.}
	\label{fig:I2Iframework}
\end{figure*}

\section{A. Detailed Experimental Settings}
The experiments were conducted on a workstation equipped with an NVIDIA RTX 4090 GPU (24 GB VRAM), with approximately 20 GB of memory used during training. The batch size was set to 8 for all experiments. The system ran on the Windows operating system, and the code was developed and executed using the PyCharm IDE. The implementation was based on PyTorch, with the deep learning framework version specified as torch==1.13.0+cu116, which includes CUDA 11.6 support.

\subsection{1. Screen-Camera Dataset Capture Directions}
To ensure diversity in viewpoint and lighting conditions, we capture 100 Screen-Camera (SC) images per direction for each screen-camera pair. The 9 directions include:
\begin{itemize}
\item Center (no tilt)
\item Top, Bottom, Left, Right (orthogonal angles)
\item Top-left 45°, Top-right 45°, Bottom-left 45°, Bottom-right 45° (diagonal angles)
\end{itemize}

These angles reflect common deviations in SC and simulate realistic user behaviors. The shooting distance varies between 20-30 cm to approximate typical smartphone usage. Furthermore, screen devices cover refresh rates of 60 Hz, 144 Hz, and 240 Hz, enabling analysis of how different temporal scanning strategies affect SC-induced noise patterns. This setup provides a representative coverage of real-world SC conditions without enumerating extreme or rare edge cases.

\subsection{2. Hyperparameter Selection}

Our hyperparameter choices are primarily inspired by Blur2Blur~\cite{pham2024blur2blur}, a representative work in unsupervised image restoration. Following their empirical insights, we set the loss weights as:
\[
\lambda_G = 1.0, \quad \lambda_{\text{grad}} = 0.005.
\]
These values yield stable performance across our experiments and demonstrate good transferability across different screen-camera pairs. While fine-tuning is possible, we find this setting sufficiently robust without the need for task-specific adjustments. A more detailed ablation study on these hyperparameters will be considered in future work.

\subsection{3. The Design of the Image-to-image Network}
For the design of the image-to-image network, we adopt MIMO-UNet \cite{fan2021rethinking}, which addresses the deblurring task through a single encoder-decoder structure, thus avoiding the high computational burden associated with multi-level subnetwork stacking in traditional coarse-to-fine methods. Specifically, the multi-output design of the decoder enables the approximation of progressive image transformation effects, reducing memory usage and computational costs, thus facilitating lightweight integration with other watermarking frameworks. By incorporating the Multi-Input Single Encoder (MISE) and Asymmetric Feature Fusion (AFF) modules, MIMO-UNet enables multi-scale feature extraction and fusion within a single network structure, effectively preserving key information across different scales and enhancing the generation of noise details. Additionally, to alleviate the mode collapse problem associated with learning SC noise and to generate a diverse set of realistic noisy images, we follow the approach suggested in \cite{zhu2017toward} and inject latent codes \( z \) into the generator during training. The modified MIMO-UNet for noise injection is shown in Figure~\ref{fig:I2Iframework}. For a more detailed description of the framework, please refer to \cite{fan2021rethinking}.

\section{B. Comparison of Robustness Under Different Shooting Distances}
To evaluate the robustness of our proposed S2R at different distances, we design several experiments to approximate SC scenarios at various distances. In the experiments, we ensure that the camera is always positioned perpendicular to the screen to eliminate the effects of shooting angles. Specifically, we select five shooting distances: 20 cm, 25 cm, 30 cm, 35 cm, and 40 cm. For each distance, we employ the same device pair for SC, and the captured images are analyzed to assess the watermark extraction performance.

\begin{table*}[!ht]
	\centering

		\begin{tabular}{ccccccc}
			\toprule[1pt]
			\multicolumn{2}{c}{\multirow{2}[2]{*}{Angle=0°}} & \multicolumn{5}{c}{\hspace{4pt}BER (\%)} \\ \cmidrule(l){3-7} 
			\multicolumn{2}{c}{}                          & \hspace{3pt}20 cm   & 25 cm   & 30 cm   & 35 cm   & 40 cm  \\ \midrule
			\multirow{4}{*}{\hspace{3pt}Method}     & StegaStamp \cite{tancik2020stegastamp}     & \hspace{3pt}2.9  & 3.9  & 4.6  & 4.7  & 4.4 \\
			& PIMoG \cite{fang2022pimog}          & \hspace{3pt}1.5  & 1.4  & 3.3  & 3.2  & 2.6 \\
			& SSDS \cite{li2024screen}             & \hspace{3pt}2.4  & 2.7  & 2.1  & 2.7  & 4.1 \\
			& S2R (SIM+LEA)    & \hspace{3pt}\textbf{1.2}  & \textbf{1.1}  & \textbf{2.1}  & \textbf{2.5}  & \textbf{2.2} \\ \bottomrule[1pt]
	\end{tabular}
        \caption{Comparison of Bit Accuracy for Extracted Watermark Message under Different Shooting Distances.\label{tab:distance}}
\end{table*}

The experimental results, as illustrated in Table~\ref{tab:distance}, clearly demonstrate that S2R outperforms other methods in terms of watermark extraction accuracy at a variety of shooting distances. The increased accuracy can be attributed to the innovative integration of mathematical modeling and deep learning techniques employed by S2R. This combination allows the model to adapt more effectively to and learn from the diverse noise patterns encountered in real-world imaging scenarios.

\section{C. Histogram Comparison between Approximate Noise Images under Different Methods}
\begin{figure*}[!t]
	\centering
	\includegraphics[width=0.9\linewidth]{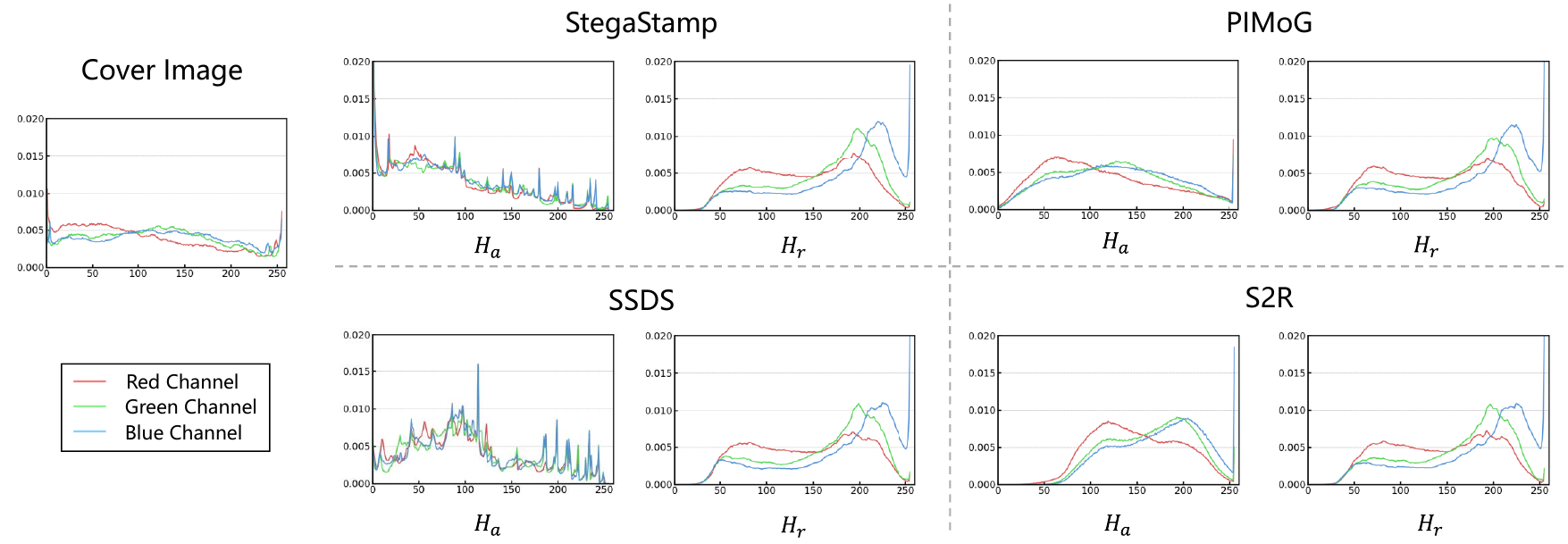}
	\caption{Comparison between approximate noise images under different schemes. Where \( H_r \) represents the histogram of real SC Image, and \( H_a \) represents the histogram of approximate noise Images. For all histograms, the horizontal axis represents pixel intensity ranging from 0 to 255, while the vertical axis represents the average normalized frequency ranging from 0 to 1.}
	\label{fig:hist}
\end{figure*}

As shown in Figure~\ref{fig:hist}, StegaStamp \cite{tancik2020stegastamp} produces extremely high peaks in the region of low pixel intensity. The PIMoG \cite{fang2022pimog} histogram curve is closer to the actual one, but it lacks depth due to simplified noise. SSDS \cite{li2024screen} has a certain level of complexity, but forms a large number of sharp peaks. On the other hand, S2R has a clear advantage in the pixel histogram distribution, successfully approximating subtle noise variations in real SC scenes.

\section{D. Additional Generalization Experiments}

To complement the main paper’s evaluation on device-level generalization, we conduct two additional experiments to assess the robustness of our framework under content distribution shift and heterogeneous training settings.

\subsection{1. Cross-Dataset Generalization}
We retain the original training setup (i.e., both the S2R module and the watermarking network train on COCO images), and test the model on two unseen datasets: ImageNet \cite{deng2009imagenet} and Mirflickr \cite{huiskes2008mir}. These test sets are constructed by applying the same SC capture process to the new content sources, ensuring consistent degradation patterns while varying semantic and texture statistics.

\begin{table}[!ht]
	\centering

    \setlength{\tabcolsep}{1mm}
		\begin{tabular}{lccc}
			\toprule[1pt]
			Test Dataset & BER (\%) & PSNR (dB) & SSIM \\ \midrule
			COCO (original test set) & 1.7 & 41.62 & 0.964 \\
			ImageNet                 & 1.7 & 41.78 & 0.967 \\
			Mirflickr                & 1.9 & 41.82 & 0.966 \\
			\bottomrule[1pt]
	\end{tabular}
        \caption{Generalization across unseen content domains. Both S2R and the watermarking network are trained on COCO; tested on different datasets.\label{tab:gen_content}}
\end{table}

As shown in Table \ref{tab:gen_content}, the results show that the model achieves nearly identical performance across unseen content domains. In fact, slight improvements in perceptual quality metrics (PSNR/SSIM) are observed, indicating strong robustness under domain shift. We attribute this to the training objective focusing primarily on learning SC distortions rather than content semantics, enabling the model to generalize across diverse styles and texture patterns.

\subsection{2. Cross-Source Training for S2R and Watermarking}
To further test the modularity of our framework, we train the S2R module on datasets different from the one used to train the watermarking encoder. The test set is fixed to Mirflickr, while the watermarking network is always trained on COCO.

\begin{table}[!ht]
	\centering
		\setlength{\tabcolsep}{1mm}
        \begin{tabular}{lccc}
			\toprule[1pt]
			S2R Training Source & BER (\%) & PSNR (dB) & SSIM \\ \midrule
			COCO (original set) & 1.9 & 41.82 & 0.966 \\
			ImageNet            & 2.3 & 41.80 & 0.952 \\
			Mirflickr           & 2.0 & 41.54 & 0.954 \\
			\bottomrule[1pt]
	\end{tabular}
        \caption{Cross-dataset training: S2R and watermarking network trained on different datasets (tested on Mirflickr).\label{tab:gen_crossdomain}}
\end{table}

As shown in Table~\ref{tab:gen_crossdomain}, the results confirm that the framework remains robust even when the S2R and watermarking modules are trained on different datasets. BER remains low across all settings. A minor SSIM drop is observed under domain mismatch, likely due to differences in image statistics affecting watermark embedding behavior.

These findings demonstrate that our system is resilient across devices, content types, and modular training setups, supporting its practicality in real-world deployments with varying training data and application scenarios.

\section{E. Correlation Between Noise Fidelity and Watermarking Robustness}
To further support our claim that improved noise modeling contributes to enhanced watermarking performance, we conduct an additional experiment to quantify the relationship between the fidelity of noise approximation and downstream robustness.

We train the S2R network and extract intermediate versions of the learned noise layer at multiple training stages (after 10, 20, 50, 100, and 300 epochs). Each intermediate noise model is then used to train a full watermarking pipeline from scratch under identical settings. The resulting watermarking models are evaluated on the COCO test set using three metrics: BER, PSNR, and SSIM.

\begin{table}[ht]
\centering
\setlength{\tabcolsep}{1mm}  
\begin{tabular}{cccc}
\toprule
S2R Training Epochs & BER (\%) & PSNR (dB) & SSIM \\
\midrule
10   & 3.0 & 41.23 & 0.947 \\
20   & 3.1 & 41.52 & 0.943 \\
50   & 2.9 & 41.22 & 0.957 \\
100  & 2.3 & 41.62 & 0.961 \\
300  & 2.1 & 41.72 & 0.963 \\
\bottomrule
\end{tabular}
\caption{Impact of S2R training epochs on watermarking performance.}
\label{tab:noise_fidelity}
\end{table}

As shown in Table~\ref{tab:noise_fidelity}, we observe a consistent trend: as the S2R noise model becomes more realistic with training (reflected by increasing PSNR and SSIM), the BER of the watermarking system decreases accordingly. This empirical evidence reinforces the hypothesis that higher fidelity in noise approximation leads to stronger robustness. The results establish a clearer correlation between noise realism and watermarking performance, directly addressing the concern regarding the lack of quantitative coupling between these factors.

\section{F. Limitations and Potential Future Improvements}

To further examine the robustness of the proposed watermarking framework, we analyze its performance under diverse SC conditions that reflect real-world distortions. These include geometric transformations, illumination imbalances, and partial occlusions, all of which are challenging to simulate precisely in the training domain.

\begin{table}[ht]
\centering
\begin{tabular}{lc}
\toprule
Capture Condition & BER (\%) \\
\midrule
Angle = 0°, Distance = 30cm       & 1.6 \\
Angle = +80°, Distance = 30cm     & 30.0 \\
Angle = –80°, Distance = 30cm     & 26.0 \\
Distance = 100cm, Angle = 0°      & 3.6 \\
Local light spot                  & 2.5 \\
Dark screen                       & 30.0 \\
Partial crop (retain center 75\%) & 50.0 \\
\bottomrule
\end{tabular}
\caption{Watermark decoding BER under representative SC conditions.}
\label{tab:ber_conditions}
\end{table}

The results indicate that the model maintains low BER under conditions such as standard viewing angles, increased viewing distance, and localized lighting artifacts. These perturbations are typically low-frequency and well captured by the simulation process. In contrast, substantial performance degradation is observed under extreme viewing angles (±80°) and global underexposure (dark screen), which introduce complex geometric and photometric changes that are underrepresented in the simulated domain $\mathcal{C}$. The most pronounced failure occurs under partial cropping, where the loss of spatial context leads to a BER of 50.0\%. This suggests that the current model lacks robustness to occlusions, likely due to the absence of such distortions during training.

These findings highlight both the strengths and limitations of the proposed approach. While the model generalizes effectively to a range of natural perturbations, its sensitivity to severe domain shifts underscores the need for broader data augmentation strategies. Future work may consider integrating hard-to-simulate distortions such as extreme perspective transformations and spatial occlusions into the training process to further enhance robustness.

\section{G. Resolution-Agnostic Watermarking by Resolution Scaling}

The proposed watermarking framework is trained and evaluated on fixed-resolution images of size $128 \times 128$, as commonly adopted in recent SC watermarking works. However, in real-world applications, SC content often appears at diverse resolutions. To bridge this gap and enable resolution-agnostic deployment, we adopt the reversible resolution scaling strategy proposed in \cite{bui2023trustmark}, as detailed in Algorithm~\ref{alg:scaling}. This technique allows any watermarking model trained at a fixed resolution to operate on arbitrary input sizes during inference, without requiring retraining. Importantly, the scaling strategy ensures that both the visual content and the spatial integrity of the embedded watermark are preserved. This allows the watermarking model, which is trained at a fixed resolution, to process high-resolution images during deployment without modification.

We incorporate this resolution scaling in the verification phase of the S2R watermarking framework, allowing encoded images to be processed at any resolutions. This eliminates the constraint of fixed-resolution watermarking and facilitates broader applicability in practical scenarios.

To support resolution-agnostic watermarking, we adopt the scaling procedure described in Algorithm~\ref{alg:scaling}, which transforms arbitrary-resolution inputs into the encoder's native resolution and reconstructs high-resolution outputs while preserving the embedded watermark.

Table~\ref{tab:res} reports the performance of S2R across a wide range of resolutions. As the results show, watermark quality remains high (PSNR$>$43 dB, SSIM$>$0.96) with consistently low Bit Error Rates (BER), even at resolutions up to 1080$\times$1920. These findings validate the effectiveness of resolution-aware adaptation and highlight the robustness of our method under real-world deployment settings.

\begin{table}[ht]
\centering
\caption{Watermarking performance of S2R across different resolutions using resolution scaling.}
\label{tab:res}
\begin{tabular}{lccc}
\toprule
\textbf{Resolution} & \textbf{PSNR (dB)} & \textbf{SSIM} & \textbf{BER (\%)} \\
\midrule
128$\times$128    & 42.27 & 0.962 & 2.1 \\
256$\times$256    & 44.52 & 0.974 & 2.5 \\
400$\times$400    & 44.11 & 0.978 & 2.9 \\
480$\times$854    & 44.01 & 0.983 & 2.6 \\
720$\times$1080   & 43.93 & 0.985 & 1.2 \\
1080$\times$1920  & 43.92 & 0.988 & 1.3 \\
\bottomrule
\end{tabular}
\end{table}

\begin{algorithm}[ht]
\caption{Resolution Scaling}
\label{alg:scaling}
\begin{algorithmic}[1]
\STATE \textbf{Input:} Input image $x_o$, binary watermark $w$
\STATE \textbf{Output:} Watermarked image $x_w$
\STATE \textbf{Model:} Watermark encoder $E(\cdot)$ trained at resolution $u \times v$
\STATE $h, w \leftarrow \text{Size}(x_o)$
\STATE $x_o \leftarrow x_o / 127.5 - 1$ \hfill // normalize to range $[-1, 1]$
\STATE $x'_o \leftarrow \text{interpolate}(x_o, (u, v))$
\STATE $r' \leftarrow E(x'_o) - x'_o$ \hfill // residual image
\STATE $r \leftarrow \text{interpolate}(r', (h, w))$
\STATE $x_w \leftarrow \text{clamp}(x_o + r, -1, 1)$
\STATE $x_w \leftarrow x_w \times 127.5 + 127.5$
\end{algorithmic}
\end{algorithm}

\section{H. Additional Visualization Examples}
To provide a deeper understanding of the proposed S2R framework and its performance, we include additional visualizations in the supplementary material:

Figure \ref{fig:unsupervise}: This figure illustrates the process of converting simulated noise into real SC noise using an unsupervised learning approach within the S2R framework. It highlights how the model effectively bridges the gap between mathematical modeling and real-world distortions.

\begin{figure*}[!htb]
	\centering
	\includegraphics[width=\linewidth]{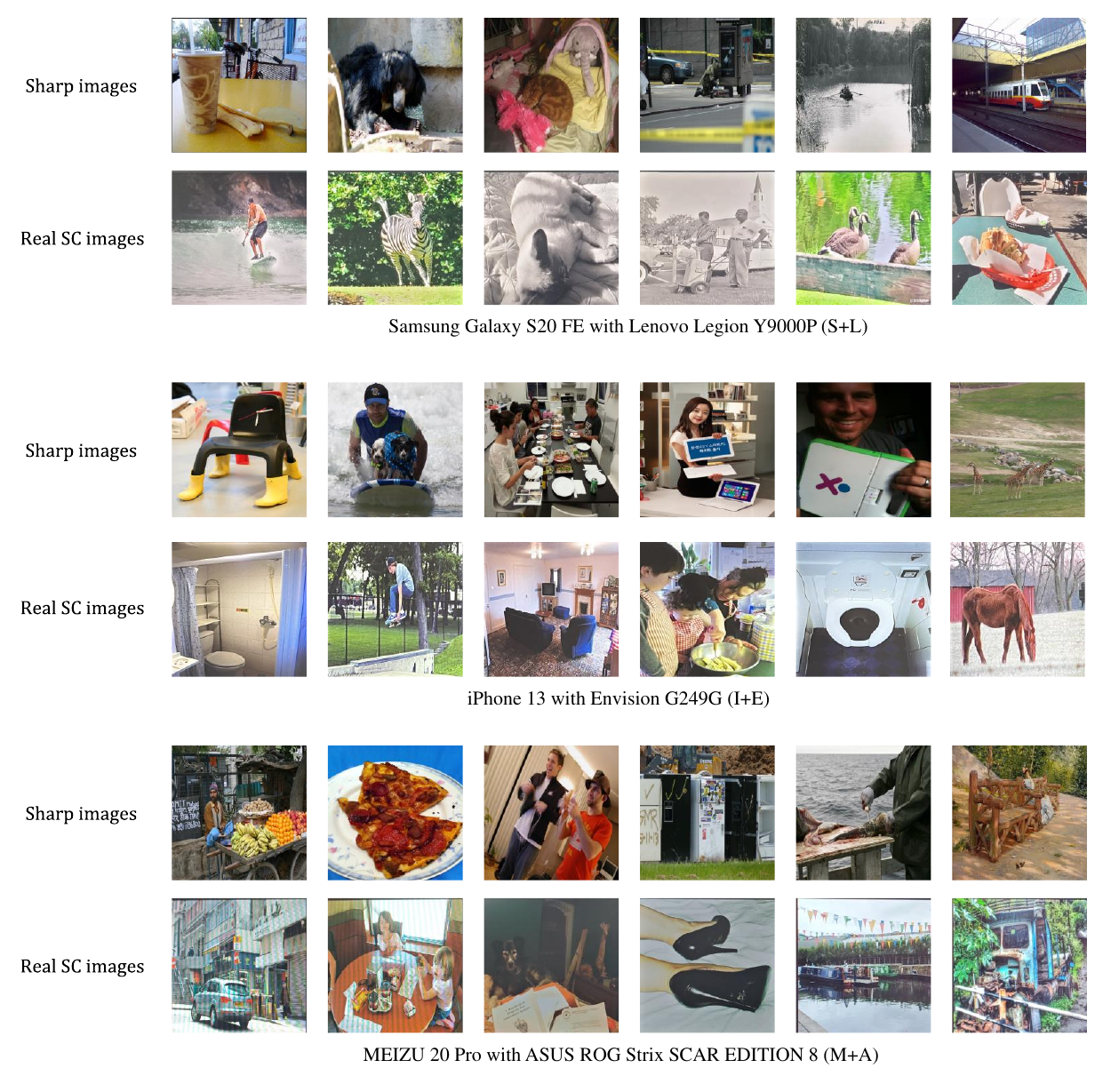}
	\caption{Visualization of COCO datasets and images captured from different devices.}
	\label{fig:datasets}
\end{figure*}

Figure \ref{fig:datasets}: This figure presents a visualization of images from the COCO dataset and those captured from different devices, showcasing the diversity of training and testing data. This diversity is essential for improving the generalization capability of the model across various real-world scenarios.

\begin{figure*}[!htb]
	\centering
	\includegraphics[width=\linewidth]{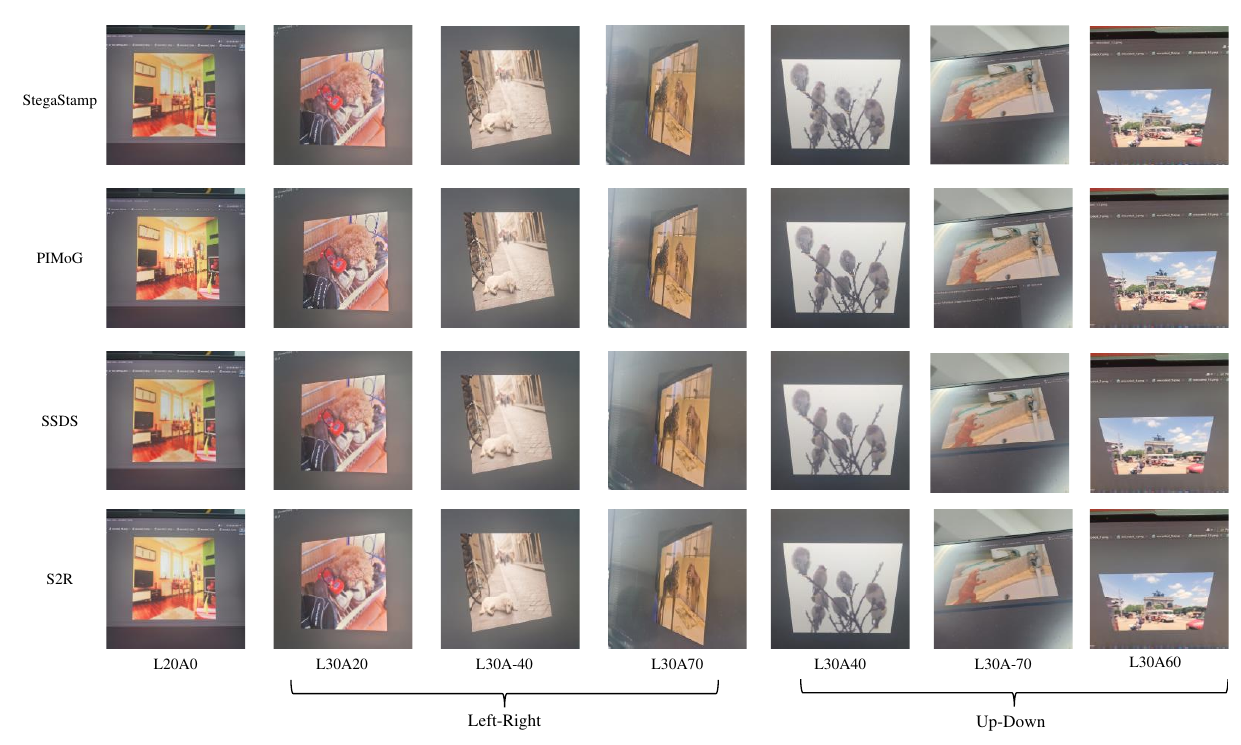}
	\caption{Real SC encoded images at different shooting angles and distances.}
	\label{fig:sc-real}
\end{figure*}

Figure \ref{fig:sc-real}: This figure displays real SC encoded images captured at different shooting angles and distances, demonstrating how varying capture conditions affect watermark robustness. These examples validate the ability of S2R to maintain high performance under diverse real-world conditions.

These figures collectively provide visual evidence supporting the effectiveness, robustness, and adaptability of S2R in real SC scenarios.

\begin{figure*}[!htb]
	\centering
	\includegraphics[width=0.7\linewidth]{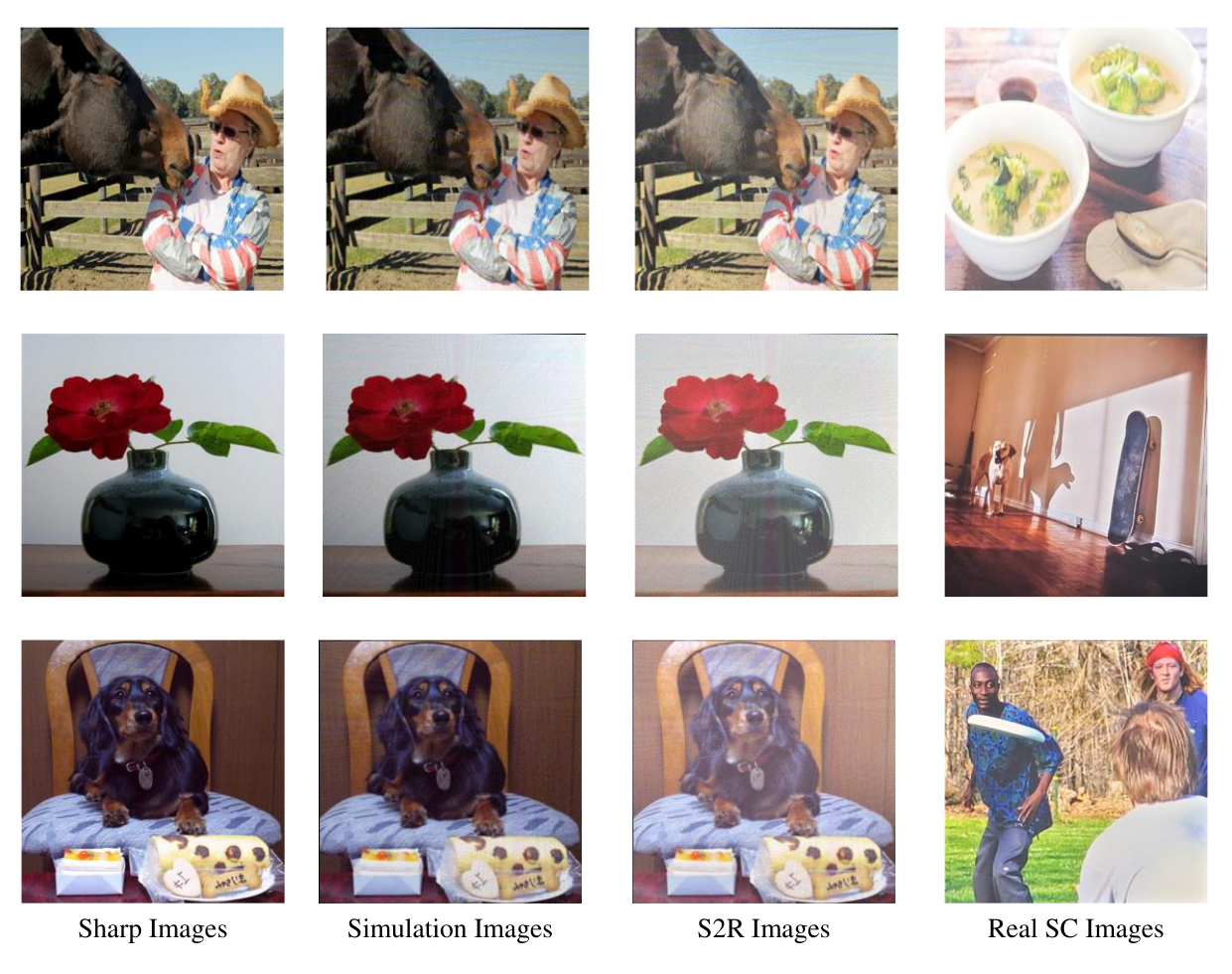}
	\caption{Visualization of the S2R converting simulated noise into real noise through unsupervised learning.}
	\label{fig:unsupervise}
\end{figure*}

\end{document}